\documentclass[10pt,twocolumn,letterpaper]{article}

\usepackage{cvpr}
\usepackage{times}
\usepackage{epsfig}
\usepackage{graphicx}
\usepackage{amsmath,bm}
\usepackage{amssymb}
\usepackage{makecell}
\usepackage{booktabs}
\usepackage{multirow}
\usepackage{color}
\usepackage[svgnames,table]{xcolor}
\usepackage{comment}

\usepackage[pagebackref=true,breaklinks=true,letterpaper=true,colorlinks,bookmarks=false]{hyperref}

\cvprfinalcopy %

\def\cvprPaperID{204} %

\def\V2L{\textit{V2L}}
\def\CNN{\mbox{CNN}}
\def\Att{{V_{att}}}
\def\ModParms{{\boldsymbol \theta}}

\ifcvprfinal\fi
\begin{document}

\title{What Value Do Explicit High Level Concepts Have \\in Vision to Language Problems?}

\author{Qi Wu, Chunhua Shen, Lingqiao Liu, Anthony Dick, Anton van den Hengel\\
School of Computer Science,
The University of Adelaide, Australia \\
{\tt\small \{qi.wu01,chunhua.shen,lingqiao.liu,anthony.dick,anton.vandenhengel\}@adelaide.edu.au}}

\maketitle
\thispagestyle{empty}

\begin{abstract}
Much recent progress in Vision-to-Language (\V2L) problems has been achieved through a combination of Convolutional Neural Networks (CNNs) and Recurrent Neural Networks (RNNs). This approach does not explicitly represent high-level semantic concepts, but rather seeks to progress directly from image features to text. In this paper we investigate whether this direct approach succeeds due to, or despite, the fact that it avoids the explicit representation of high-level information. We propose a method of incorporating high-level concepts into the successful CNN-RNN approach, and show that it achieves a significant improvement on the state-of-the-art in both image captioning and visual question answering. We also show that the same mechanism can be used to introduce external semantic information and that doing so further improves performance. We achieve the best reported results on both image captioning and VQA on several benchmark datasets, and provide an analysis of the value of explicit high-level concepts in \V2L problems.
\end{abstract}

\vspace{-10pt}
\section{Introduction}

Vision-to-Language problems present a particular challenge in Computer Vision because they require translation between two different forms of information.  In this sense the problem is similar to that of machine translation between languages.
In machine language translation there have been a series of results showing that good performance can be achieved without developing a higher-level model of the state of the world. In~\cite{bahdanau2014neural,cho2014learning,sutskever2014sequence}, for instance, a source sentence is transformed into a fixed-length vector representation by an `encoder' RNN, which in turn is used as the initial hidden state of a `decoder' RNN that generates the target sentence.  
\begin{figure}[t]
  \centering
  \includegraphics[width=0.95\linewidth]{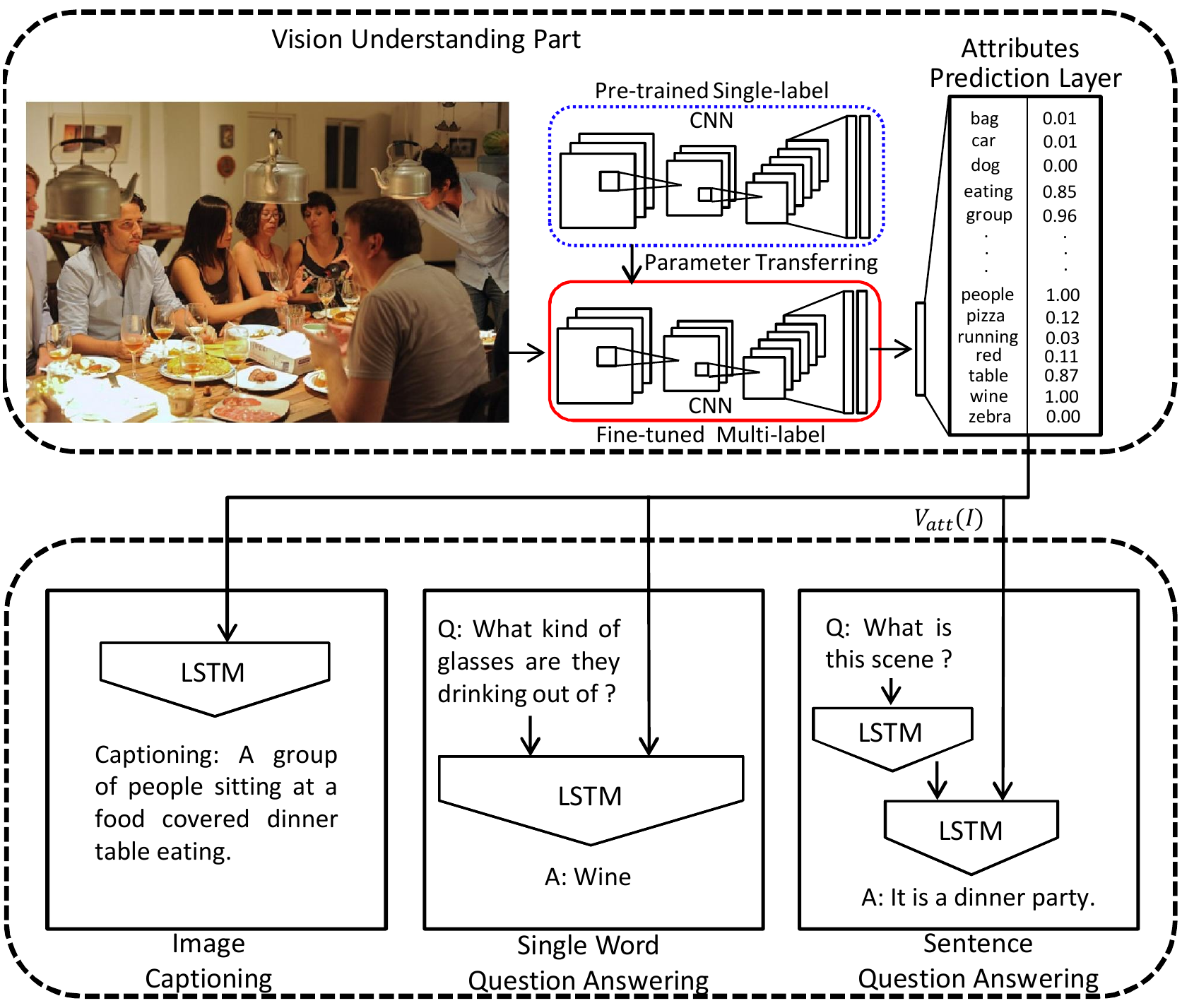}\\
  \caption{Our attribute based \textit{V2L} framework. The image analysis module learns a mapping between an image and the semantic attributes through a CNN. The language module learns a mapping from the attributes vector to a sequence of words using an LSTM. 
	}
	\label{img:frame}
  \vspace{-15pt}
\end{figure}

Despite the supposed equivalence between an image and 1000 words, the manner in which information is represented in each data form could hardly be more different. Human language is designed specifically so as to communicate information between humans, whereas even the most carefully composed image is the culmination of a complex set of physical processes over which humans have little control.  Given the differences between these two forms of information, it seems surprising that methods inspired by machine language translation have been so successful.  These RNN-based methods which translate directly from image features to text, without developing a high-level model of the state of the world, represent the current state of the art for key Vision-to-Language (\V2L) problems, such as image captioning and visual question answering.  

This approach is reflected in many recent successful works on image captioning, such as ~\cite{Chen2015CVPRMind,donahue2014long,karpathy2014deep,mao2014deep,vinyals2014show,yao2015describing}.  
Current state-of-the-art 
captioning methods use a CNN as an image `encoder' to produce a fixed-length vector representation~\cite{krizhevsky2012imagenet,lecun1998gradient,simonyan2014very,szegedy2014going}, which is then fed into the `decoder' RNN to generate a caption.

Visual Question Answering (VQA) is a more recent challenge than image captioning.  In this \V2L problem an image and a free-form, open-ended question about the image are presented to the method which is required to produce a suitable answer~\cite{antol2015vqa}. Same as image captioning, the current state of the art in VQA~\cite{gao2015you,malinowski2015ask,ren2015image} relies on passing CNN features to an RNN language model.

Our main contribution is to consider the question: 
\textit{what value do explicit high level concepts have in V2L problems?}  
That is, given that significant performance improvements have been achieved by moving to models which directly pass from image features to text, should we give up on  high-level concepts in \V2L altogether?
We investigate particularly the impact that adding high-level information to the CNN-RNN framework has upon performance.  We do this by inserting an explicit representation of attributes of the scene which are meaningful to humans.  
Each semantic attribute corresponds to a word mined from the training image descriptions, and represents higher-level knowledge about the content of the image.
A CNN-based classifier is trained for each attribute, and the set of attribute likelihoods for an image forms a high-level representation of image content.  An RNN is then trained to generate captions, or answer questions, on the basis of the likelihoods.

Our second contribution is a fully trainable attribute based neural network that can be applied to multiple \textit{V2L} problems  which yields significantly better performance than current state-of-the-art approaches. For example, in the Microsoft COCO Captioning Challenge, we produce a BLEU-1 score of 0.73, which is the state of the art on the leaderboard at the time of writing. Our final model also provides the state-of-the-art performance on several recently released VQA datasets. For instance, our system yields a WUPS@0.9 score of 71.15, compared with the current state of the art of 66.78, on the Toronto COCO-QA single word question answering dataset. On the VQA (test-standard), an open-answer task dataset, our method achieves 55.84\% accuracy, while the baseline is 54.06\%. Moreover, with an expansion from image-sourced attributes to knowledge-sourced through WordNet (see Section \ref{subsec:att_exp}), we further improve the accuracy to 57.62\%.
\section{Related Work}
\label{Related_Work}

\paragraph{Image Captioning} The problem of annotating images with natural language at the scene level has long been studied in both computer vision and natural language processing. Hodosh \etal~\cite{hodosh2013framing} proposed to frame sentence-based image annotation as the task of ranking a given pool of captions. Similarly,~\cite{gong2014improving,jia2011learning,ordonez2011im2text} posed the task as a retrieval problem, but based on co-embedding of images and text in the same space. Recently, Socher \etal \cite{socher2014grounded} used neural networks to co-embed image and sentences together and Karpathy \etal \cite{karpathy2014deep} co-embedded image crops and sub-sentences. Neither attempted to generate novel captions.

Attributes have been used in many image captioning methods to fill the gaps in predetermined caption templates.
Farhadi \etal~\cite{farhadi2010every}, for instance, used detections to infer a triplet of scene elements which is converted to text using a template. Li \etal \cite{li2011composing} composed image descriptions given computer vision based inputs such as detected objects, modifiers and locations using web-scale $n$-grams. %
A more sophisticated CRF-based method that uses attribute detections beyond triplets was proposed by Kulkarni \etal ~\cite{kulkarni2013babytalk}. The advantage of template-based methods is that the resulting captions are more likely to be grammatically correct. The drawback is that they still rely on hard-coded visual concepts and suffer the implied limits on the variety of the output. Instead of using fixed templates, more powerful language models based on language parsing have been developed, such as~ \cite{aker2010generating,kuznetsova2012collective,kuznetsova2014treetalk,mitchell2012midge}. 

Fang \etal \cite{fang2014captions} won the 2015 COCO Captioning Challenge with an approach that is similar to ours in as much as it applies a visual concept (i.e., attribute) detection process before generating sentences. They first learned $1000$ independent detectors for visual words based on a multi-instance learning framework and then used a maximum entropy language model conditioned on the set of visually detected words directly to generate captions. 
Differently, our visual attributes act as a high-level semantic representation for image content which is fed into an LSTM which generates target sentences based on a much larger word vocabulary. More importantly, the success of their model relies on a re-scoring process from a joint image-text embedding space. To what extent the high-level concepts help in image captioning (and other \V2L tasks) is not discussed in their work. Instead, this is the main focus of this paper.

In contrast to the aforementioned two-stage methods, the recent dominant trend in \V2L is to use an architecture which connects a CNN to an RNN to learn the mapping from images to sentences directly. Mao \etal \cite{mao2014deep}, for instance, proposed a multimodal RNN (m-RNN) to estimate the probability distribution of the next word given previous words and the deep CNN feature of an image at each time step. Similarly, Kiros \etal \cite{kiros2014unifying} constructed a joint multimodal embedding space using a powerful deep CNN model and an LSTM that encodes text. Karpathy \etal \cite{Karpathy2014deepvs} also proposed a multimodal RNN generative model, but in contrast to \cite{mao2014deep}, their RNN is conditioned on the image information only at the first time step. Vinyals \etal \cite{vinyals2014show} combined deep CNNs for image classification with an LSTM for sequence modeling, to create a single network that generates descriptions of images. Chen \etal \cite{Chen2015CVPRMind} learned a bi-directional mapping between images and their sentence-based descriptions using RNN. Xu \etal \cite{xu2015show} proposed a model based on visual attention, as well as You \etal~\cite{Att_CVPR_2016}. Jia \etal \cite{jia2015guilding} applied additional retrieved sentences to guide the LSTM in generating captions. Devlin \etal~\cite{devlin2015language} combined both maximum entropy (ME) language model and RNN to generate captions.

Interestingly, this end-to-end CNN-RNN approach ignores the image-to-word mapping which was an essential step in many of the previous image captioning systems detailed above~\cite{farhadi2010every,kulkarni2013babytalk,li2011composing,yang2011corpus}. The CNN-RNN approach has the advantage that it is able to generate a wider variety of captions, can be trained end-to-end, and outperforms the previous approach on the benchmarks. It is not clear, however, what the impact of bypassing the intermediate high-level representation is, and particularly to what extent the RNN language model might be compensating. Donahue \etal \cite{donahue2014long} described an experiment, for example, using tags and CRF models as a mid-layer representation for video to generate descriptions, but it was designed to prove that LSTM outperforms an SMT-based approach \cite{rohrbach2013translating}. It remains unclear whether the mid-layer representation or the LSTM leads to the success. Our paper provides several well-designed experiments to answer this question.

We thus here show not only a method for introducing a high-level representation into the CNN-RNN framework, and that doing so improves performance, but we also investigate the value of high-level information more broadly in \V2L tasks.  This is of critical importance at this time because \V2L has a long way to go, particularly in the generality of the images and text it is applicable to.  

\vspace{-10pt}
\paragraph{Visual Question Answering}
Visual question answering is one of the more challenging, and interesting, \V2L tasks as it requires answering previously unseen questions about image content~\cite{antol2015vqa, gao2015you,ma2015learning,malinowski2014multi,malinowski2015hard,malinowski2015ask,ren2015image,zhu2015building}. This is as opposed to the vast majority of challenges in Computer Vision in which the question is specified long before the program is written.  Both Gao \etal \cite{gao2015you} and Malinowski \etal \cite{malinowski2015ask} used RNNs to encode the question and output the answer. %
Ren \etal \cite{ren2015image} focused on questions with a single-word answer and formulated the task as a classification problem using an LSTM, and released a single-word answer dataset (Toronto COCO-QA). Ma \etal \cite{ma2015learning} used CNNs to both extract image features and sentence features, and fuse the features together with a multi-modal CNN. 
Antol \etal \cite{antol2015vqa} proposed a large-scale open-ended VQA dataset based on COCO, which is called VQA. They also provided several baseline methods which combined both image features (CNN extracted) and question features (LSTM extracted) to obtain a single embedding and further built a MLP (Multi-Layer Perceptron) to obtain a distribution over answers. %
\vspace{-5pt}
\section{An Attribute-based \V2L Model}
Our approach is summarized in Figure \ref{img:frame}. The model includes an image analysis part and a language generation part. In the image analysis part, we first use supervised learning to predict a set of attributes, based on words commonly found in image captions. We solve this as a multi-label classification problem and train a corresponding deep CNN by minimizing an element-wise logistic loss function. Secondly, a fixed length vector $\Att(I)$ is created for each image $I$, whose length is the size of the attribute set. Each dimension of the vector contains the prediction probability for a particular attribute. In the language generation part, we apply an LSTM-based sentence generator. Our attribute vector $\Att(I)$ is used as an input to this LSTM. For different tasks, we have different  language models. For image captioning, we follow \cite{vinyals2014show} to generate sentences from an  LSTM; for single-word question answering, as in~\cite{ren2015image}, we use the LSTM as a classifier 
providing a likelihood for each potential answer; for open-ended question answering, we use an encoder LSTM to encode questions while the second LSTM decoder uses the attribute vector $\Att(I)$ to generate a sentence based answer. A baseline model is also implemented for each of the three tasks. In the baseline model, as in~\cite{gao2015you,ren2015image,vinyals2014show} we use a pre-trained CNN to extract image features $\CNN(I)$ which are fed into the LSTM directly. For the sake of completeness a fine-tuned version of this approach is also implemented. The baseline method is used as a counterpart to verify the effectiveness of the intermediate attribute prediction layer for each task.

\subsection{The Attribute Predictor}
\label{subsec:Attributes_Predictor}
We first build an attributes vocabulary regardless of the final tasks (\ie image captioning, VQA).
Unlike~\cite{kulkarni2013babytalk,yang2011corpus}, that use a vocabulary from separate hand-labeled training data, our semantic attributes are extracted from training captions and can be any part of speech, including object names (nouns), motions (verbs) or properties (adjectives). The direct use of captions guarantees that the most salient attributes for an image set are extracted. We use the $c$ most common words in the training captions to determine the attribute vocabulary. %
In contrast to~\cite{fang2014captions}, our vocabulary is not tense or plurality sensitive (done manually), for instance, \texttt{`ride'} and \texttt{`riding'} are classified as the same semantic attribute, similarly \texttt{`bag'} and \texttt{`bags'}. This significantly decreases the size of our attribute vocabulary. We finally obtain a vocabulary with 256 attributes. %
Our attributes represent a set of high-level semantic constructs, the totality of which the LSTM then attempts to represent in sentence form. Generating a sentence from a vector of attribute likelihoods exploits a much larger set of candidate words which are learned separately (see Section \ref{subsec:Language_Generator} for more details).

Given this attribute vocabulary, we can associate each image with a set of attributes according to its captions. We then wish to predict the attributes given a test image. Because we do not have ground truth bounding boxes for attributes, 
we cannot train a detector for each using the standard approach.
 Fang \etal~\cite{fang2014captions} solved a similar problem using a Multiple Instance Learning framework~\cite{zhang2005multiple} to detect visual words from images. 
Motivated by the relatively small number of times that each word appears in a caption, we instead
treat this as a multi-label classification problem. 
To address the concern that some attributes may only apply to image sub-regions, we follow Wei \etal~\cite{wei2014cnn} in designing a region-based multi-label classification framework. %

\begin{figure}[t]
  \centering
  \includegraphics[width=1\linewidth]{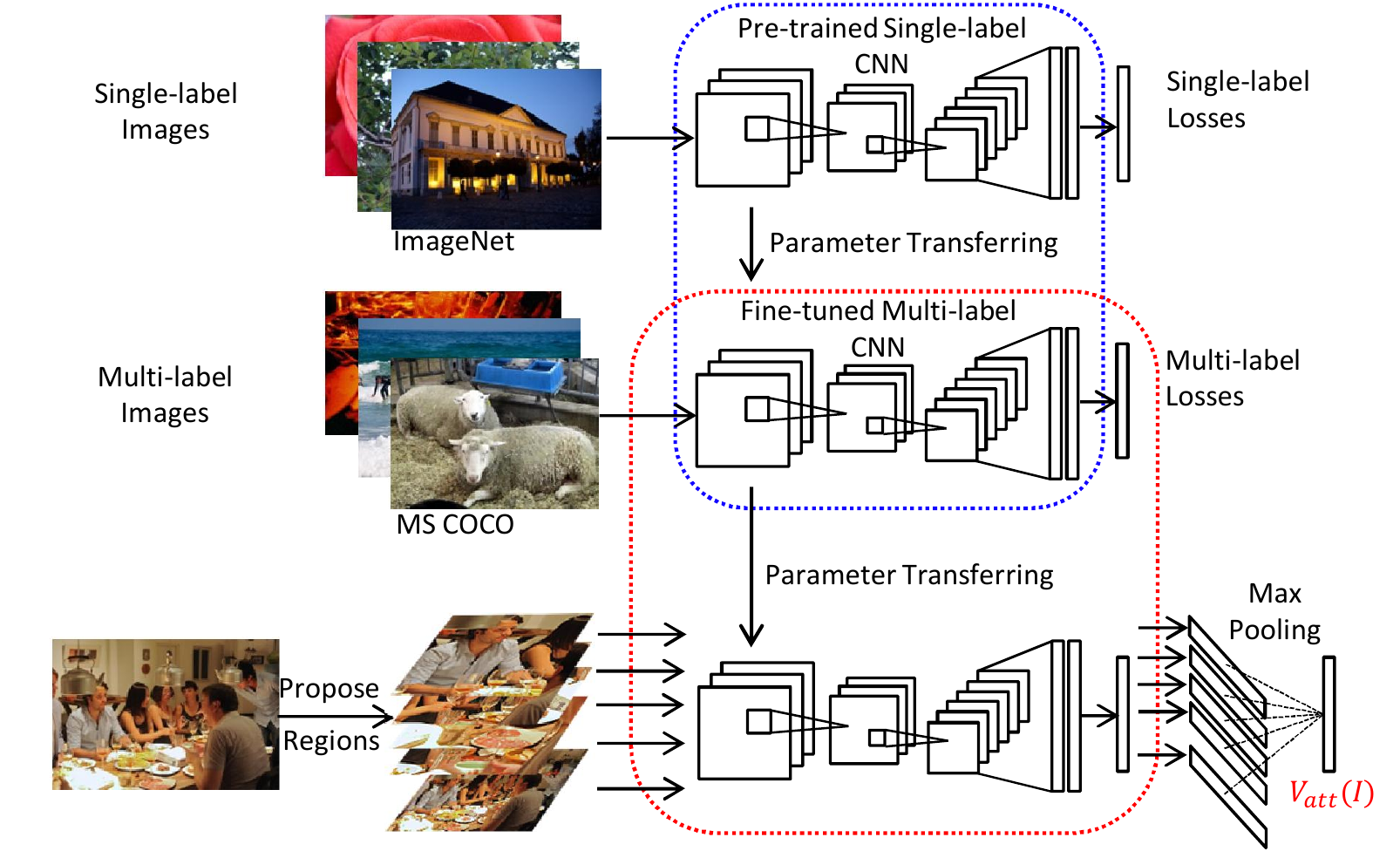}\\
  \caption{Attribute prediction CNN: the model is initialized from VggNet \cite{simonyan2014very} pre-trained on ImageNet. The model is then fine-tuned on the target multi-label dataset. Given a test image, a set of proposal regions are selected and passed to the shared CNN, and finally the CNN outputs from different proposals are aggregated with max pooling to produce the final multi-label prediction, which gives us the high-level image representation, $\Att(I)$}
  \label{img:attributes}
  \vspace{-14pt}
\end{figure}

Figure \ref{img:attributes} summarizes the attribute prediction network. In contrast to~\cite{wei2014cnn}, which uses AlexNet \cite{krizhevsky2012imagenet} as the initialization of the shared CNN, we use the more powerful VggNet \cite{simonyan2014very} pre-trained on ImageNet~\cite{deng2009imagenet}. This model has been widely used in image captioning tasks~\cite{Chen2015CVPRMind,fang2014captions,Karpathy2014deepvs,mao2014deep}. The shared CNN is then fine-tuned on the target multi-label dataset (our image-attribute training data). In this step, the output of the last fully-connected layer is fed into a $c$-way softmax. %
The $c=256$ here represents the attribute vocabulary size. In contrast to \cite{wei2014cnn} who employs the squared loss, we find that element-wise logistic loss function performs better. Suppose that there are $N$ training examples and $\bm{y_i}=[y_{i1}, y_{i2},... , y_{ic}]$ is the label vector of the $i^{th}$ image, where $y_{ij}=1$ if the image is annotated with attribute $j$, and $y_{ij}=0$ otherwise. If the predictive probability vector is $\bm{p_i}=[p_{i1}, p_{i2},... , p_{ic}]$, then the cost function to be minimized is 
\vspace{-2pt}
\begin{equation}
 J=\frac{1}{N}\sum_{i=1}^{N}\sum_{j=1}^{c}\log(1+\exp(-y_{ij}p_{ij}))
 \vspace{-2pt}
\end{equation}
During the fine-tuning process, the parameters of the last fully connected layer (i.e. the attribute prediction layer) are initialized with a Xavier initialization \cite{glorot2010understanding}. The learning rates of  `\texttt{fc6}' and `\texttt{fc7}' of the VggNet are initialized as 0.001 and the last fully connected layer is initialized as 0.01. All the other layers are fixed during training. We executed 40 epochs in total and decreased the learning rate to one tenth of the current rate for each layer after 10 epochs. The momentum is set to 0.9. The dropout rate is set to 0.5.

To predict attributes based on regions, we first extract hundreds of proposal windows from an image. However, considering the computational inefficiency of deep CNNs, the number of proposals processed needs to be small. Similar to \cite{wei2014cnn}, we first apply the normalized cut algorithm to group the proposal bounding boxes into $m$ clusters based on the IoU scores matrix. The top $k$ hypotheses in terms of the predictive scores reported by the proposal generation algorithm are kept and fed into the shared CNN. In contrast to \cite{wei2014cnn}, we also include the whole image in the hypothesis group. As a result, there are $mk+1$ hypotheses for each image. We set $m=10,k=5$ in all  experiments. We use Multiscale Combinatorial Grouping (MCG)~\cite{PABMM2015} for the proposal generation. Finally, a cross hypothesis max-pooling is applied to integrate the outputs into a single prediction vector $\Att(I)$.

\subsection{Language Generator}
\label{subsec:Language_Generator}

\begin{figure}[t]
  \centering
  \includegraphics[width=1\linewidth]{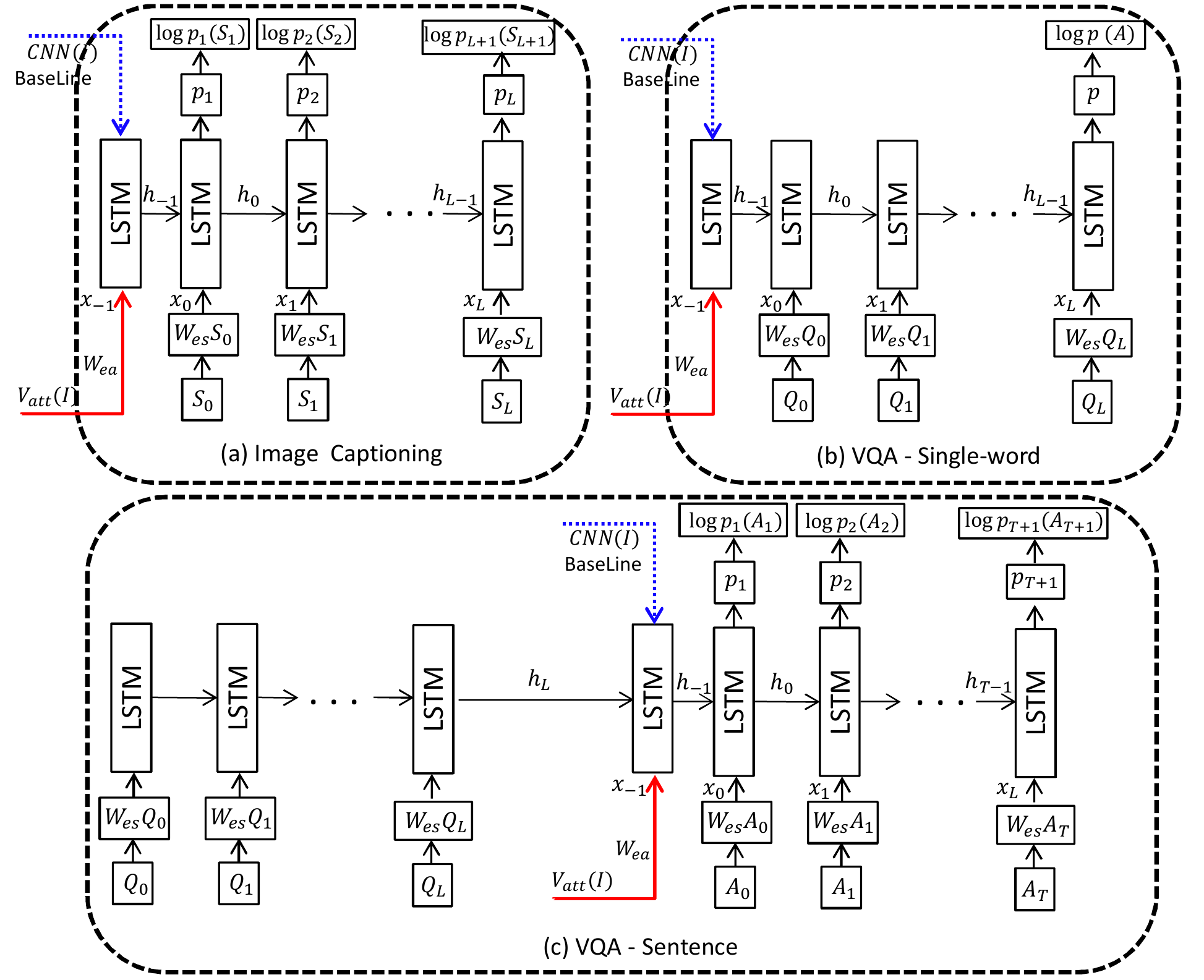}\\
  \caption{Language generators for different types of tasks: (a) Image Captioning, (b) VQA-single word, (c) VQA-sentence. \textcolor[rgb]{1.00,0.00,0.00}{red} arrow indicates our attributes input $\Att(I)$ while \textcolor[rgb]{0.00,0.00,1.00}{blue} dash arrow shows the baseline method input $\CNN(I)$.}
  \label{img:language_models}
  \vspace{-10pt}
\end{figure}

Similar to~\cite{Karpathy2014deepvs,mao2014deep,vinyals2014show}, we propose to train a language generation model by maximizing the probability of the correct description given the image. However, rather than using image features directly as in typically the case, we use the semantic attribute prediction probability $\Att(I)$ from the previous section as the input. Suppose that $\{S_1,...,S_L\}$ is a sequence of words. The log-likelihood of the words given their context words and the corresponding image can be written as:
\vspace{-3pt}
\begin{equation}
    \log p(S|\Att(I))=\sum_{t=1}^L \log p(S_{t}|S_{1:t-1},\Att(I))
\vspace{-2pt}
\end{equation}
where $p(S_t|S_{1:t-1},\Att(I))$ is the probability of generating the word $S_t$ given attribute vector $\Att(I)$ and previous words $S_{1:t-1}$. We employ the LSTM \cite{hochreiter1997long}, a particular form of RNN, to model this. %
See Figure \ref{img:language_models} for different language generators designed for multiple \V2L tasks.

\vspace{-7pt}
\paragraph{Image Captioning Model}
\label{para:image_captioning}
The LSTM model for image captioning is trained in an unrolled form. More formally, the LSTM takes the attributes vector $\Att(I)$ and a sequence of words $S=(S_0,...,S_L,S_{L+1})$, where $S_0$ is a special start word and $S_{L+1}$ is a special END token. Each word has been represented as a one-hot vector $S_t$ of dimension equal to the size of words dictionary. The words dictionaries are built based on words that occur at least 5 times in the training set, %
which lead to 8791 words on  MS COCO datasets. 
Note it is different from the semantic attributes vocabulary $\mathcal{V}_{att}$. The training procedure is as following (see Figure \ref{img:language_models} (a)) : At time step $t=-1$, we set $x_{-1}=W_{ea}\Att(I)$ and $h_{initial}=\vec{0}$, where $W_{ea}$ is the learnable attributes embedding weights. The LSTM memory state is initialized to the range $(-0.1,0.1)$ with a uniform distribution. This gives us an initial LSTM hidden state $h_{-1}$ which can be used in the next time step. From $t=0$ to $t=L$, we set $x_t=W_{es}S_t$ and the hidden state $h_{t-1}$ is given by the previous step, where $W_{es}$ is the learnable word embedding weights. The probability distribution $p_{t+1}$ over all words is then computed by the LSTM feed-forward process. Finally, on the last step when $S_{L+1}$ represents the last word, the target label is set to the END token.

Our training objective is to learn parameters $W_{ea}$, $W_{es}$ and all parameters in LSTM by minimizing the following cost function:
\vspace{-5pt}
\begin{eqnarray}
    \mathcal{C}&=&-\frac{1}{N}\sum_{i=1}^N\log p(S^{(i)}|\Att(I^{(i)}))+\lambda_\ModParms\cdot||\ModParms||_2^2 \\
    &=&-\frac{1}{N}\sum_{i=1}^N\sum_{t=1}^{L^{(i)}+1}\log p_t(S_t^{(i)})+\lambda_\ModParms\cdot||\ModParms||_2^2
    \vspace{-3pt}
\end{eqnarray}
where $N$ is the number of training examples and $L^{(i)}$ is the length of the sentence for the $i$-th training example. $p_t(S_t^{(i)})$ corresponds to the activation of the Softmax layer in the LSTM model for the $i$-th input and $\ModParms$ represents model parameters, $\lambda_\ModParms\cdot||\ModParms||_2^2$ is a regularization term. We use SGD with mini-batches of 100 image-sentence pairs. The attributes embedding size, word embedding size and hidden state size are all set to 256 in all the experiments. The learning rate is set to 0.001 and clip gradients is 5. The dropout rate is set to 0.5.

\vspace{-10pt}
\paragraph{Question Answering Model}

For question answering, a triplet $\{\Att(I),\{Q_1,...,Q_L\},\{A_1,...,A_T\}\}$ is given, whereas $L$ and $T$ is the length of the question and answer, separately. We define it to be a single-word answering problem when $T=1$ and a sentence-based problem if $T>1$.

For the single-word answering problem, the LSTM takes the attributes score vector $\Att(I)$ and a sequence of input words of the question $Q=(Q_1,...,Q_L)$. The feed-forward process is the same as image captioning, except that an END token is not required anymore. Instead, we use the word generated by the last word of the question as the predicted answer (see Figure \ref{img:language_models} (b)). Hence, the cost function is $\mathcal{C}=-\frac{1}{N}\sum_{i=1}^N\log p(A^{(i)})+\lambda_\ModParms\cdot||\ModParms||_2^2$, where $N$ is the number of training examples. $\log p(A^{(i)})$ is the log-probability distribution over all candidate answers that is computed by the last LSTM cell, given the previous hidden state and the last word of question~$Q_L$.

For the sentence-based question answering, we have a question encoding LSTM and an answer decoding LSTM. However, different from Gao \etal \cite{gao2015you} using two separates LSTMs for question and answer, weights between our encoding and decoding LSTMs are shared. The information stored in the LSTM memory cells of the last word in the question is treated as the representation of the sentence. And its hidden state will be used as the initial state of the answering LSTM part. Moreover, different from \cite{gao2015you,malinowski2015ask,ren2015image} who use CNN features directly, we use our attributes representations $\Att(I)$ as the input for decoding LSTM (see Figure \ref{img:language_models} (c)). The cost function of sentence-based question answering is $\mathcal{C}=-\frac{1}{N}\sum_{i=1}^N\sum_{t=1}^{T^{(i)}+1}\log p_t(A_t^{(i)})+\lambda_\ModParms\cdot||\ModParms||_2^2$, where $T^{(i)}+1$ is the length of the answer plus one END token for the $i$-th training example. According to training configuration, the learning rate is set to 0.0005 and other parameters are same as image captioning configuration. 
\section{Image Captioning}
\subsection{Dataset}

There are several datasets which consist of images and sentences describing them in English. We mainly report results on the popular Microsoft COCO~\cite{lin2014microsoft} dataset. Results on Flickr8k \cite{hodosh2013framing} and Flickr30k \cite{young2014image} can be found in the supplementary material. MS COCO contains 123,287 images, and each image is annotated with 5 sentences. Because most previous work in image captioning \cite{donahue2014long,fang2014captions,Karpathy2014deepvs,mao2014deep,vinyals2014show,xu2015show} is not evaluated on the official test split of MS COCO, for fair comparison, we report results with the widely used publicly available splits in the work of \cite{Karpathy2014deepvs}, which use 5000 images for validation, and 5000 for testing. We further tested on the actual MS COCO test set consisting of 40775 images (human captions for this split are not available publicly), and evaluated them on the COCO evaluation server.

\vspace{-3pt}
\subsection{Evaluation}
\vspace{-5pt}
\paragraph{Metrics}
We report results with the frequently used BLEU metric and sentence perplexity ($\mathcal{PPL}$). BLEU \cite{papineni2002bleu} scores are originally designed for automatic machine translation where they measure the fraction of $n$-grams (up to 4-gram) that are in common between a hypothesis and a reference or set of references. Here we compare against 5 references. Perplexity is a standard measure for evaluating language models which measures how many bits on average would be needed to encode each word given the language model, so a low $\mathcal{PPL}$ means a better language model. Additionally, we evaluate our model based on the metrics METEOR \cite{banerjee2005meteor}, and CIDEr \cite{vedantam2014cider}. All scores (except $\mathcal{PPL}$) are computed with the \texttt{coco-evaluation} code~\cite{chen2015microsoft}.

\vspace{-10pt}
\paragraph{Baselines}
To verify the effectiveness of our attribute representation, we provide a baseline method. The baseline framework is the same as that proposed in section \ref{para:image_captioning}, except that the attributes vector $\Att(I)$ is replaced by the last hidden layer of CNN directly (see the blue arrow in Figure~\ref{img:language_models}). Various CNN architectures are applied in the baseline method to extract image features, such as VggNet\cite{simonyan2014very} and GoogLeNet\cite{szegedy2014going}. For the \textbf{VNet+LSTM}, we use the second fully connected layer (\texttt{fc7}), which has 4096 dimensions. In \textbf{VNet-PCA+LSTM}, PCA is applied to decrease the feature dimension from 4096 to 1000. For the \textbf{GNet+LSTM}, we use the GoogleNet model provided in the Caffe Model Zoo \cite{jia2014caffe} and the last average pooling layer is employed, which is a 1024-d vector. \textbf{VNet+ft+LSTM} applies a VggNet that has been fine-tuned on the target dataset, based on the task of image-attributes classification.

\vspace{-5pt}
\paragraph{Our Approaches}
We evaluate several variants of our approach: \textbf{Att-GT+LSTM} models use ground-truth attributes as the input while \textbf{Att-CNN+LSTM} uses the attributes vector $\Att(I)$ predicted by the attributes prediction network in section \ref{subsec:Attributes_Predictor}. We also evaluate an approach \textbf{Att-SVM+LSTM} with linear SVM ($C=1$) predicted attributes vector. SVM classifiers are trained to divide positive attributes from those negatives given an image-attributes correspondence. We use the second fully connected layer of the fine-tuned VggNet to feed the SVM. To infer the sentence given an input image, we use Beam Search, which iteratively considers the set of $b$ best sentences up to time $t$ as candidates to generate sentences at time $t+1$, and only keeps the best $b$ results. We set the $b$ as 5.

\begin{table}[t]
\scriptsize
\begin{center}
\resizebox{\linewidth}{!}{%
  \begin{tabular}{ c c c c c c c|c}
    \Xhline{2\arrayrulewidth}
    \textbf{State-of-art} &B-1 & B-2 & B-3 & B-4 & M & C & $\mathcal{P}$ \\ \hline
    NeuralTalk \cite{Karpathy2014deepvs}& 0.63 & 0.45 & 0.32 & 0.23 & 0.20& 0.66&- \\
    Mind's Eye \cite{Chen2015CVPRMind}& -&-&-&0.19&0.20&-&11.60 \\
    NIC \cite{vinyals2014show}& - & - & - &0.28&0.24&0.86&- \\
    LRCN \cite{donahue2014long}&0.67&0.49&0.35&0.25&-&-&- \\
    Mao et al.\cite{mao2014deep}&0.67&0.49&0.34&0.24&-&-&13.60 \\
    Jia et al.\cite{jia2015guilding}&0.67&0.49&0.36&0.26&0.23&0.81&- \\
    MSR \cite{fang2014captions}&-&-&-&0.26&0.24&-&18.10\\
    Xu et al.\cite{xu2015show}&0.72&0.50&0.36&0.25&0.23&-&- \\
    Jin et al.\cite{jin2015aligning}&0.70&0.52&0.38&0.28&0.24&0.84&- \\
    \Xhline{2\arrayrulewidth}
    \textbf{Baseline-\textit{CNN(I)}} &  &  &  & & & & \\ \hline
    VNet+LSTM & 0.61 & 0.42 & 0.28 & 0.19 &0.19 &0.56&13.58  \\
    VNet-PCA+LSTM & 0.62 & 0.43 & 0.29 & 0.19 &0.20 &0.60&13.02 \\
    GNet+LSTM &0.60  &0.40  & 0.26 &0.17  & 0.19&0.55&14.01  \\
    VNet+ft+LSTM & 0.68 & 0.50 & 0.37 & 0.25 & 0.22&0.73& 13.29 \\
    \Xhline{2\arrayrulewidth}
    \textbf{Ours-$\Att(I)$} & & & & & &&\\ \hline
    \cellcolor[rgb]{0.7,0.7,0.7}{Att-GT+LSTM\textsuperscript{$\ddagger$}} & \cellcolor[rgb]{0.7,0.7,0.7}{0.80} & \cellcolor[rgb]{0.7,0.7,0.7}{0.64} & \cellcolor[rgb]{0.7,0.7,0.7}{0.50} & \cellcolor[rgb]{0.7,0.7,0.7}{0.40} &\cellcolor[rgb]{0.7,0.7,0.7}{0.28} & \cellcolor[rgb]{0.7,0.7,0.7}{1.07}&\cellcolor[rgb]{0.7,0.7,0.7}{9.60} \\
    Att-SVM+LSTM & 0.69 & 0.52 & 0.38  &0.28 & 0.23& 0.82&12.62  \\
    Att-CNN+LSTM & \textbf{0.74} & \textbf{0.56} & \textbf{0.42}  & \textbf{0.31} & \textbf{0.26}& \textbf{0.94}& \textbf{10.49} \\ \hline
  \end{tabular}}
      \vspace{1pt}
      \caption{BLEU-1,2,3,4, METEOR, CIDEr and $\mathcal{PPL}$ metrics compared with other state-of-the-art methods and our baseline on MS COCO dataset. $\ddagger$ indicates ground truth attributes labels are used, which (in \colorbox[rgb]{0.7,0.7,0.7}{gray}) will not participate in rankings.} %
      \label{tabcoco}
      \vspace{-15pt}
\end{center}
\end{table}

\vspace{-5pt}
\paragraph{Results} Table \ref{tabcoco} reports image captioning results on the COCO. It is not surprising that \textbf{Att-GT+LSTM} model performs best, since ground truth attributes labels are used. We report the results just to show the advances of adding an intermediate image-to-word mapping stage. Ideally, if we are able to train a strong attributes predictor which gives us a good enough estimation of attributes, we could obtain an outstanding improvement comparing with both baselines and state-of-the-arts. Indeed, apart from using ground truth attributes, our \textbf{Attributes-CNN+LSTM} models generate the best results over all evaluation metrics. Especially comparing with baselines, which do not contain an attributes prediction layer, our final models bring significant improvements, nearly 15\% for B-1 and 30\% for CIDEr on average. \textbf{VNet+ft+LSTM} model performs better than other baselines because of the fine-tuning on the target dataset. However, they do not perform as good as our attributes-based models. \textbf{Att-SVM+LSTM} under-performs \textbf{Att-CNN+LSTM} means our region-based attributes prediction network performs better than the SVM classifier. Our final model also outperforms current state of the arts listed in tables. We also evaluate an approach that combines CNN features and attributes vector together as the input of the LSTM, but we find this approach (B-1=0.71) is not as good as using attributes vector alone in the same setting. In any case, above experiments show that an intermediate image-to-words stage (i.e. attributes prediction layer) brings us significant improvements. Results on Flickr8k and Flickr30k can be found in the supplementary material, as well as some qualitative results.

\begin{table}[t!]
\scriptsize
\begin{center}
\resizebox{\linewidth}{!}{
  \begin{tabular}{ c c c c c c c c }
    \Xhline{2\arrayrulewidth}
    COCO-TEST & B-1 & B-2 & B-3&B-4 & M & R &CIDEr\\ \hline
    \textbf{5-Refs} &&&&&&& \\\hline
    Ours&\textbf{0.73}&\textbf{0.56}&\textbf{0.41}&\textbf{0.31}&\textbf{0.25}&\textbf{0.53}&\textbf{0.92}\\
    Human&0.66&0.47&0.32&0.22&\textbf{0.25}&0.48&0.85\\
    MSR \cite{fang2014captions}&0.70&0.53&0.39&0.29&\textbf{0.25}&0.52&0.91\\
    m-RNN \cite{mao2014deep}&0.68&0.51&0.37&0.27&0.23&0.50&0.79\\
    LRCN \cite{donahue2014long}&0.70&0.53&0.38&0.28&0.24&0.52&0.87\\
    \Xhline{2\arrayrulewidth}
    \textbf{40-Refs} &&&&&&& \\\hline
    Ours&\textbf{0.89}&\textbf{0.80}&\textbf{0.69}&\textbf{0.58}&0.33&\textbf{0.67}&\textbf{0.93}\\
    Human&0.88&0.74&0.63&0.47&\textbf{0.34}&0.63&0.91\\
    MSR \cite{fang2014captions}&0.88&0.79&0.68&0.57&0.33&0.66&\textbf{0.93}\\
    m-RNN \cite{mao2014deep}&0.87&0.76&0.64&0.53&0.30&0.64&0.79\\
    LRCN \cite{donahue2014long}&0.87&0.77&0.65&0.53&0.32&0.66&0.89\\
    \Xhline{2\arrayrulewidth}
  \end{tabular}}
  \vspace{1pt}
  \caption{COCO evaluation server results. M and R stands for METEOR and ROUGE-L.
  Results using 5 references and 40 references captions are both shown. We only list the comparison results that have been officially published in the corresponding references.}
  \label{tab3}
  \vspace{-15pt}
\end{center}
\end{table}

We further generated captions for the images in the COCO test set containing 40,775 images and evaluated them on the COCO evaluation server. These results are shown in Table \ref{tab3}. We achieve 0.73 on B-1, and surpass human performances on 13 of the 14 metrics reported. We are the best results on 3 evaluations metrics (B-1,2,3) on the server leaderboard at the time of writing this paper. We also achieve the top-5 ranking on the other evaluation metrics.

Table \ref{tab3-2} summarizes some properties of recurrent layers employed in some recent RNN-based methods.  We achieve state-of-the-art using a relatively small dimensional visual input feature and recurrent layer. Lower dimension of visual input and RNN normally means less parameters in the RNN training stage, as well as lower computation cost.

\begin{table}[t]
\scriptsize
\begin{center}
\resizebox{\linewidth}{!}{
  \begin{tabular}{ c c c c c c}
    \Xhline{2\arrayrulewidth}
      & Ours & NIC\cite{vinyals2014show} & LRCN\cite{donahue2014long} & m-RNN\cite{mao2014deep} & NeuralTalk\cite{Karpathy2014deepvs}\\ \hline
    VIS Input Dim&256&1000&1000&4096&4096\\
    RNN Dim&256&512&1000$\times4$&256&300-600\\
    \Xhline{2\arrayrulewidth}
  \end{tabular}}
  \vspace{1pt}
  \caption{Visual feature input dimension and properties of RNN. Our visual features has been encoded as a 256-d attributes score vector while other models need higher dimensional features to feed to RNN. According to the unit size of RNN, we achieve state-of-the-art using a relatively small dimensional recurrent layer. 
  }
  \label{tab3-2}
  \vspace{-20pt}
\end{center}
\end{table}

\section{Visual Question Answering}
\subsection{Dataset}
We report VQA results on two recently publicly available visual question answering datasets, both are created based on MS COCO. %
Toronto COCO-QA dataset \cite{ren2015image} contains four types of questions, specifically the object, number, color and location. The answers are all single-word. We use this dataset to examine our single-word question answering model. VQA \cite{antol2015vqa} is a much larger dataset which contains 614,163 questions. These questions and answers are sentence-based and open-ended. The training and testing split follows COCO official split, which contains 82,783 training images, 40,504 validation images and 81,434 test images, each has 3 questions and 10 answers. 
We use the official test split for our testing.

\subsection{Evaluation}
Our experiments in question answering are designed to verify the effectiveness of introducing the intermediate attribute layer. Hence, apart from listing several state of art methods, we focus on comparing with a baseline method, which only uses the second fully connected layer (fc7) of the VggNet (and a fine-tuned VggNet) as the input.

Table \ref{tab4} reports results on the Toronto COCO-QA dataset, within which all answers are a single-word. Besides the accuracy value (the proportion of correct answered testing questions to the total testing questions), the Wu-Palmer similarity (WUPS)~\cite{wu1994verbs} is also used to measure the performance of different models. The WUPS calculates the similarity between two words based on the similarity between their common subsequence in the taxonomy tree. If the similarity between two words is greater than a threshold then the candidate answer is assumed to be right. We follow~\cite{ma2015learning,ren2015image} in setting the threshold as 0.9 and 0.0. \textbf{GUESS} is a simple baseline to predict the 
most common answer from the training set
based on the question type. The modes are `cat', `two', `white', and `room' for the four types of questions. 
\textbf{VIS+BOW}~\cite{ren2015image} performs multinomial logistic regression based on image features and a BOW vector obtained by summing all the word vectors of the question. \textbf{VIS+LSTM}~\cite{ren2015image} has one LSTM to encode the image and question, while \textbf{2-VIS+BLSTM} has two image feature input points, at the start and the end of the sentences. Ma \textit{et al.}~\cite{ma2015learning} encoded both images and questions by CNN. 
From the Table \ref{tab4}, we clearly see that our attribute-based model outperforms the baselines and all state-of-the-art methods by a significant degree, which proves the effectiveness of our attribute-based representation for \textit{V2L} tasks.

\begin{table}[t]
\scriptsize
\begin{center}
\resizebox{\linewidth}{!}{
    \begin{tabular}{ c|c c c }
    \Xhline{2\arrayrulewidth}
    \textbf{Toronto COCO-QA} &Acc & WUPS@0.9 & WUPS@0.0 \\ \hline
    GUESS\cite{ren2015image}&6.65&17.42&73.44 \\
    VIS+BOW\cite{ren2015image}&55.92&66.78&88.99 \\
    VIS+LSTM\cite{ren2015image}&53.31&63.91&88.25 \\
    2-VIS+BLSTM\cite{ren2015image}&55.09&65.34&88.64 \\
    Ma et al.\cite{ma2015learning}&54.94&65.36&88.58\\
    \Xhline{2\arrayrulewidth}
    \textbf{BaseLine} &  &   &   \\ \hline
    VggNet-LSTM & 50.73 & 60.37 & 87.48 \\
    VggNet+ft-LSTM & 58.34 & 67.32 & 89.13 \\
    \Xhline{2\arrayrulewidth}
    \textbf{Our-Proposal} &  &   &   \\ \hline
    \cellcolor[rgb]{0.7,0.7,0.7}{Att-GT+LSTM}\textsuperscript{$\ddagger$}&\cellcolor[rgb]{0.7,0.7,0.7}{67.66}&\cellcolor[rgb]{0.7,0.7,0.7}{75.76}&\cellcolor[rgb]{0.7,0.7,0.7}{93.63} \\
    Att-CNN+LSTM& \textbf{61.38}& \textbf{71.15}& \textbf{91.58} \\
     \Xhline{2\arrayrulewidth}
    \end{tabular}}
    \vspace{1pt}
    \caption{Accuracy, WUPS@0.9 and WUPS@0.0 metrics compared with other state-of-the-art methods and our baseline on the Toronto COCO-QA dataset. Each image has one question and only a single word answer is given for each. $\ddagger$~indicates that ground truth attributes labels were used, and thus that the method does not participate in rankings.}
    \label{tab4}
    \vspace{-24pt}
\end{center}
\end{table}

\begin{table}[b!]
\vspace{-13pt}
\centering
\resizebox{\linewidth}{!}{
\begin{tabular}{ccccccccc}
\Xhline{2\arrayrulewidth}
\multicolumn{1}{c|}{}         & \multicolumn{4}{c|}{Test-dev}                       & \multicolumn{4}{c}{Test-standard} \\
\multicolumn{1}{c|}{}         & All   & Y/N   & Num   & \multicolumn{1}{c|}{Others} & All    & Y/N    & Num    & Others \\ \hline
\multicolumn{1}{c|}{Q+I \cite{antol2015vqa}}   & 52.64 & 75.55 & 33.67 & \multicolumn{1}{c|}{37.37}  & -  & -  & -  & -  \\
\multicolumn{1}{c|}{LSTM Q \cite{antol2015vqa}}   & 48.76 & 78.20 & 35.68 & \multicolumn{1}{c|}{26.59}  & 48.89  & 78.12  & 34.94  & 26.99  \\
\multicolumn{1}{c|}{LSTM Q+I \cite{antol2015vqa}} & 53.74 & 78.94 & 35.24 & \multicolumn{1}{c|}{36.42}  & 54.06  & 79.01  & 35.55  & 36.80  \\
\multicolumn{1}{c|}{Human \cite{antol2015vqa}}    & -     & -     & -     & \multicolumn{1}{c|}{-}      & 83.30  & 95.77  & 83.39  & 72.67  \\ \hline
\multicolumn{1}{c|}{VNet+ft+LSTM}  & 55.03 & 78.19 & 35.47 & \multicolumn{1}{c|}{39.68}  & 55.34  & 78.10  & 35.30  & 40.27  \\
\multicolumn{1}{c|}{Att-CNN+LSTM}     & 55.57 & 78.90 & 36.11 & \multicolumn{1}{c|}{40.07}  & 55.84  & 78.73  & \textbf{36.08}  & 40.60  \\
\multicolumn{1}{c|}{Att-KB+LSTM}      & \textbf{57.46} & \textbf{79.77} & \textbf{36.79} & \multicolumn{1}{c|}{\textbf{43.10}}  & \textbf{57.62}  & \textbf{79.72}  & 36.04 & \textbf{43.44}  \\
\Xhline{2\arrayrulewidth}
\end{tabular}}
\caption{Results on test-dev and test-standard split of VQA dataset compared with \cite{antol2015vqa}.}
\label{tab5}
\end{table}
Table \ref{tab5} summarizes the results on the test split of VQA dataset. In contrast to the above single-word question answering task, here we follow \cite{antol2015vqa}, and measure performance by recording the percentage of answers in agreement with ground truth from human subjects. 
Antol \etal \cite{antol2015vqa} provided a baseline for this dataset using a \textbf{Q+I} method, which encodes the image with CNN features and questions with LSTM representation. Then they train a softmax neural network classifier with a single hidden layer and the output space is the 1000 most frequent answers in the training set. Human performance is also given in \cite{antol2015vqa} for reference.
\textbf{VNet+ft+LSTM} is the model with fine-tuned VggNet features. It is slightly less accurate than our explicit attributes based model \textbf{Att-CNN+LSTM}, but the gap is small. \textbf{LSTM Q+I} \cite{antol2015vqa} can be treated as our baseline as it uses CNN features as the input to the LSTM, while \textbf{LSTM~Q} only provides questions as the input. Our attributes based model outperforms \textbf{LSTM Q+I} nearly in all cases, especially when the answer types are `others'. Our hypothesis is that this performance increase occurs because the separately-trained attribute layer discards irrelevant image information.  This ensures that the LSTM does not interpret irrelevant variations in the expression of the text as relating to irrelevant image details, and try to learn a mapping between them. 

However, there is still a big gap between our proposed models and the human performance. After looking into details, we notice that accuracies on some question types such as `why' are very low. These kinds of questions are hard to answer because commonsense knowledge and reasoning is normally required. Zhu \textit{et al.} \cite{zhu2015building} cast a MRF model into a Knowledge Base representation to answer commonsense-related visual questions. Our semantic attribute representation 
offers hope of a solution, however, as it can be used as a key by which to source other, external information.
In the following experiment, we propose to expand our image-based attributes set to a knowledge-based attributes set through a large lexical ontology - the~WordNet.%

\subsection{Attribute Expansion using WordNet}
\label{subsec:att_exp}
WordNet \cite{miller1995wordnet} 
records a variety of relationships between words, some of which we hope to use to address the many ways of expressing the same idea in natural language.
The most frequently encoded relation is the hyponymy (such as \texttt{bed} and \texttt{bunkbed}). Meronymy represents the part-whole relation. 
Verb synsets are arranged into hierarchies (troponyms) (such as \texttt{buy}-\texttt{pay}). %
All these relationships are defined based on commonsense knowledge. %

To expand our image-sourced attributes to knowledge-sourced information, we first select candidate words from WordNet. Candidate words must fulfill two selection criteria. The first is that the word must directly linked with an arbitrary word in our attribute vocabulary $\mathcal{V}_{att}$ through the WordNet. Secondly, the candidate word must appear in at least 5 training question examples. In our experiment, given $M=256$ image-sourced attributes, we finally mined a knowledge-sourced vocabulary $\mathcal{V}_{kb}$ with $N=9762$ words, and $\mathcal{V}_{kb}$ has covered all the words in $\mathcal{V}_{att}$. Then, a similarity matrix $S \in \mathbb{R}^{M\times N}$ is computed based on a pre-trained word2vec model \cite{mikolov2013distributed}, where $S_{ij}$ gives both semantic and syntactic similarity between word $i$ in $\mathcal{V}_{att}$ and word $j$ in $\mathcal{V}_{kb}$. 
Given an image $I$ and its image-sourced attribute vector $\Att(I)=(v_{att}^{(1)},...,v_{att}^{(i)},...,v_{att}^{(M)})$ predicted by the attribute prediction network, the $j^{th}$ component of the knowledge-sourced attribute vector is obtained by a max-pooling operator $v_{kb}^{(j)}=\max(v_1^{(j)},...,v_i^{(j)},...,v_M^{(j)})$, where $v_{i}^{(j)}=v_{att}^{(i)}\times S_{ij}$. The final knowledge-sourced attributes vector $V_{kb}(I)=(v_{kb}^{(1)},...,v_{kb}^{(j)},...,v_{kb}^{(N)})$ will be fed into the LSTM to generate answers.

\begin{table}[t]
\scriptsize
\begin{center}
\resizebox{\linewidth}{!}{
    \begin{tabular}{ l c c c }
    \Xhline{2\arrayrulewidth}
    Question-Type &Vgg+LSTM&Att-CNN+LSTM & Att-KB+LSTM\\ \hline
    why&3.04&7.77&9.88 \\
    what kind&24.15&41.22&45.23\\
    which&31.28&36.60&37.28\\
    is the&71.49&73.22&74.59\\
    is this&73.00&75.26&76.63\\
    \Xhline{2\arrayrulewidth}
    \end{tabular}}
    \vspace{1pt}
    \caption{Results on the open-answer task for some commonsense reasoning question types on validation split of VQA.}
    \label{tab7}
    \vspace{-25pt}
\end{center}
\end{table}

Table \ref{tab7} compares results using image-sourced attributes vs. knowledge-sourced on the validation split of VQA dataset. We gain a significant improvement in commonsense reasoning related questions. For example, on the `why'  questions, we achieve 9.88\%. 
Our hypothesis is that this reflects the fact that indexing into WordNet in this manner provides some independence as to the exact manner of expression used in the text, but also adds extra information.  In answering questions about beds and hammocks, for example, it is useful to know that both are related to sleep. The overall performance of this \textbf{Att-KB+LSTM} model on the test split of VQA can be found in the Table \ref{tab5}. Our overall result is 57.62\% accuracy, which performs better than the model of \textbf{Att-CNN+LSTM} (the model before attributes expansion) and achieves the state-of-the-art result on the VQA dataset.

\vspace{-5pt}
\section{Conclusion}
\vspace{-2pt}
We have described an investigation into the value of high level concepts in \V2L problems, motivated by the belief that
without an explicit representation of the content of an image it is very difficult to answer reason about it.
In the process we examined the effect of introducing an intermediate attribute prediction layer into the predominant CNN-LSTM framework. We implemented three attribute-based models for the tasks of image captioning, single-word question answering and sentence question answering.
We have shown that an explicit representation of image content improves \V2L performance, in all cases.
Indeed, at the time of writing this paper, our image captioning model outperforms the state of the art on several captioning datasets. Our question answering models perform best on the Toronto COCO-QA datasets, producing an accuracy of 61.38\%. It also achieves the state of the art on the VQA, at 57.62\%, which is a big improvement over the baseline. Moreover, attribute representation enables access to high-level commonsense knowledge, which is necessary for answering commonsense reasoning related questions.

{\bf Acknowledgements}
This research was in part supported by the Data to Decisions Cooperative Research Centre.

{\small
\bibliographystyle{ieee}
\bibliography{myref}
}

\end{document}